\documentclass{article}
\usepackage{ijcai23}

\usepackage{times}
\usepackage{soul}
\usepackage{url}
\usepackage[hidelinks]{hyperref}
\usepackage[utf8]{inputenc}
\usepackage[small]{caption}
\usepackage{graphicx}
\usepackage{amsmath}
\usepackage{amsthm}
\usepackage{amsfonts,amssymb}
\usepackage{booktabs}
\usepackage{algorithm}
\usepackage{algorithmic}
\usepackage[switch]{lineno}

\usepackage[caption=false,font=footnotesize,labelfont=sf,textfont=sf]{subfig}
\usepackage{multirow}
\usepackage[abs]{overpic}
\usepackage{color}
\usepackage{colortbl}
\usepackage{xcolor}
\definecolor{lightgray}{gray}{0.7}
\usepackage{epsfig}
\usepackage{cuted}
\definecolor{url_color}{RGB}{42, 83, 163}
\usepackage{eso-pic}
\usepackage{diagbox}
\definecolor{lightgray}{gray}{0.7}

\title{NeRF-Det++: Incorporating Semantic Cues and Perspective-aware Depth Supervision for Indoor Multi-View 3D Detection}

\author{
Chenxi Huang$^{1,2}$\and
Yuenan HOU$^{2*}$\and
Weicai Ye$^{1}$\and
Di Huang$^3$\and
Xiaoshui Huang$^2$\and\\
Binbin Lin$^{1*}$\and
Deng Cai$^1$\and
Wanli Ouyang$^2$\\
\affiliations
$^1$the State Key Laboratory of CAD\&CG, College of Computer Science, Zhejiang University\\
$^2$Shanghai AI Laboratory     $^3$The University of Sydney\\
\emails
\{hcx\_98, binbinlin\}@zju.edu.cn,
\{houyuenan, huangxiaoshui\}@pjlab.org.cn,\\
\{maikeyeweicai, dihuanginfo\}@gmail.com,
dengcai78@qq.com,
wanli.ouyang@sydney.edu.au
}
\begin{document}
\def\algorithmname{NeRF-Det++}

\twocolumn[{%
% \vspace{-3em}
    \renewcommand\twocolumn[1][]{#1}%
    \setlength{\tabcolsep}{0.0mm} %0
    \newcommand{\sz}{0.125}  % 0.125 0.11
    \maketitle
    \begin{center}
        \newcommand{\teaserwidth}{\textwidth}
    \vspace{-1em}
        \includegraphics[width=0.87\linewidth]{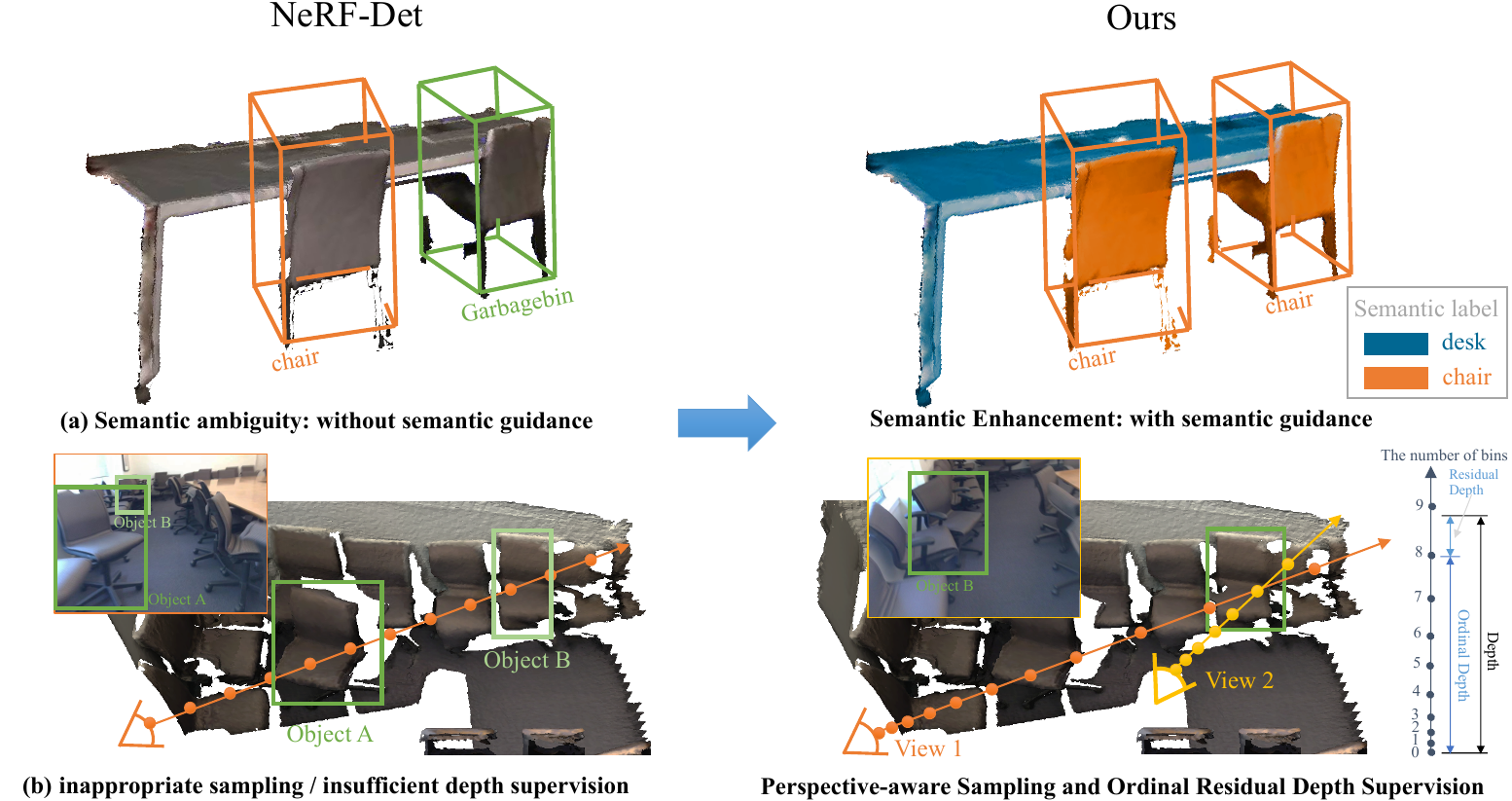}
      \vspace{-1em}
        \captionof{figure}{\textbf{The shortcomings of NeRF-Det and our corresponding solutions.}
Fig.~(a) shows the problem of semantic ambiguity. We introduce the Semantic Enhancement module that leverages semantic supervision to enhance the categorical awareness of the detector.
Fig.~(b) illustrates the limitations of the inappropriate sampling strategy and the insufficient depth supervision. 
We propose the novel Perspective-aware Sampling, which focuses more on the near, deviating from the conventional uniform sampling approach. 
Allowing different perspectives to focus on the objects that deserve more attention makes indoor multi-view 3D detection more effective. For example, we can improve the learning of object $B$ from View $2$ while allowing View $1$ to allocate more attention to its nearby objects.
Furthermore, instead of directly regressing the original depth values that are hard to optimize, we propose the Ordinal Residual Depth Supervision. 
It comprises the classification of the ordinal depth bins and the regression of the residual depth values, which is conducive to more stable depth learning.}
    \label{Fig: motivation}
    \end{center}
}]

\renewcommand{\thefootnote}{}
\footnotetext{$^*$The equal corresponding authors.}

%%%%%%%%% ABSTRACT
\begin{abstract}	
NeRF-Det has achieved impressive performance in indoor multi-view 3D detection by innovatively utilizing NeRF to enhance representation learning. Despite its notable performance, we uncover three decisive shortcomings in its current design, including \textbf{semantic ambiguity}, \textbf{inappropriate sampling}, and \textbf{insufficient utilization of depth supervision}.
To combat the aforementioned problems, we present three corresponding solutions: 1) \textbf{Semantic Enhancement}. We project the freely available 3D segmentation annotations onto the 2D plane and leverage the corresponding 2D semantic maps as the supervision signal, significantly enhancing the semantic awareness of multi-view detectors. 2) \textbf{Perspective-aware Sampling}. Instead of employing the uniform sampling strategy, we put forward the perspective-aware sampling policy that samples densely near the camera while sparsely in the distance, more effectively collecting the valuable geometric clues. 3) \textbf{Ordinal Residual Depth Supervision}. As opposed to directly regressing the depth values that are difficult to optimize, we divide the depth range of each scene into a fixed number of ordinal bins and reformulate the depth prediction as the combination of the classification of depth bins as well as the regression of the residual depth values, thereby benefiting the depth learning process. 
%supervision we put more emphasis on the nearby regions and designed the loss to concentrate on the depth learning of adjacent areas. 
The resulting algorithm, \algorithmname, has exhibited appealing performance in the ScanNetV2 and ARKITScenes datasets. Notably, in ScanNetV2, \algorithmname~outperforms the competitive NeRF-Det by \textbf{+1.9\%} in mAP$@0.25$ and \textbf{+3.5\%} in mAP$@0.50$. The code will be publicly at \url{https://github.com/mrsempress/NeRF-Detplusplus}.

\end{abstract}

%%%%%%%%% BODY TEXT
\section{Introduction}
\label{sec:intro}
%////////////////////////////////
3D object detection, which localizes and classifies the objects in the 3D space, serves as a cornerstone technique for various applications, such as augmented reality, robotics, and autonomous driving~\cite{pc_survey}. 
Multi-view 3D detection draws increasing attention from academic and industrial communities due to the wide accessibility and cheap cost of cameras and reconstruction techniques~\cite{huang2021comprehensive,multi_view_det_survey}. 
Although image-based outdoor 3D detection has been extensively studied in the past decades, the exploration of indoor multi-view 3D detection is still in its infancy. 
Indoor 3D detection is challenging due to high object density, severe occlusion, large morphological diversity, irregular spatial distributions of objects, etc.

%In outdoor scenes, objects within the same category have similar shapes and have rich geometric priors, such as their tendency to be positioned on the ground. 
%However, in indoor scenes, objects have more morphological diversity and are not always placed on the floor. They may be placed on or imbedded in other objects.
%These characteristics make indoor 3D detection more difficult than outdoor 3D detection.
%\yuenan{list the challenges of indoor 3D detection}

%we focus on indoor 3D detection given the posed imagses as input, where the objects are captured from different viewpoints.
%\yuenan{Need to highlight indoor 3D detection}

Previous efforts in indoor multi-view 3D detection have primarily focused on leveraging geometric priors, as images lack precise geometric measurement, which poses significant challenges to obtaining a comprehensive and accurate 3D representation of the surrounding environment. 
One line of methods (\emph{e.g.},~\cite{VEDet},~\cite{ImVoxelNet},~\cite{fcaf3d}) constructs geometric constraints based on the relationship among different views, and utilizes depth maps obtained through Multi-View Stereo (MVS) algorithms as supervision signals. 
Fusing these estimated depth maps effectively constructs relatively accurate representations of the 3D world. 
Another line of methods (\emph{e.g.},~\cite{PARQ}) utilizes attention~\cite{transformer} to incorporate 3D position as a geometric signal and resort to the attention mechanism to establish connections between 2D images and the 3D world, enabling the network to grasp global geometric information.

Recently, the Neural Radiance Field (NeRF)~\cite{NeRF} has emerged as a powerful tool for geometry modeling. 
Among the NeRF-based pipelines, NeRF-RPN~\cite{NeRF-RPN} and NeRF-Det~\cite{NeRF-Det} are two representatives that leverage NeRF to enhance geometry modeling and detect objects more precisely with the reinforced 3D representation. 
NeRF-RPN utilizes NeRF to extract RGB and density from sampled points and feeds the volumetric features to the perception backbone to yield the detection outputs, which causes additional costs and heavily hinges on the learned neural representation.
By leveraging the opacity field estimated by NeRF, NeRF-Det introduces no extra cost during inference and enhances the geometric awareness of the detector.

%\yuenan{This paragraph is too long. We need to shorten it. Since NeRF-Det serves as the foundation of our method, we need to include more descriptions about it.}

%\yuenan{We need one figure to show the three motivations of our method.}

Despite the impressive performance of NeRF-Det~\cite{NeRF-Det}, we identify three critical flaws in the current framework through comprehensive experiments, \emph{i.e.}, \textbf{semantic ambiguity}, \textbf{inappropriate sampling}, and \textbf{insufficient utilization of depth supervision}. 
As shown in Fig.~\ref{Fig: motivation} (a), although NeRF-Det can roughly estimate the spatial location of an object, the category is usually misclassified. 
Besides, as shown in Fig.~\ref{Fig: motivation} (b), since NeRF-Det employs uniform depth supervision, the depth loss is typically overwhelmed by the distant regions, which have few visual cues and more significant errors, making the depth learning unbalanced and ineffective.
In addition, uniform sampling adopted by NeRF-Det fails to fully use the valuable visual clues in the multiple views, weakening the benefit of the neural rendering process.
%\yuenan{A bit general. We need to be more specific.}
%While geometric priors are crucial for accurately positioning 3D bounding boxes, semantic priors play an indispensable role in classifying the contents of objects.
%Therefore, we propose a novel approach that leverages differentiable rendering and fuses diverse priors from multiple 2D views to bridge this gap and be both geometry-aware and semantic-aware.

%\yuenan{We need to highlight the role of proper depth supervision and sampling strategy. And we need more insights.}

To tackle the above-mentioned flaws, we systematically investigate the reasons and present the following solutions: 1) \textbf{Semantic Enhancement}. We project the freely available 3D segmentation annotations onto the 2D plane and leverage the corresponding 2D semantic maps as the supervision signal. The introduced semantic enhancement module leverages semantic cues to increase the categorical awareness of the detector. 2) \textbf{Perspective-aware Sampling}. The multi-view complementation motivates us to design perspective-aware sampling. As shown in Fig.~\ref{Fig: motivation} (b), an object in a distant region in one view can be located in a nearby region in another view.
By employing perspective-aware sampling, the view with richer visual cues is assigned higher learning weights, making the learning more effective. 3) \textbf{Ordinal Residual Depth Supervision}. Directly regressing the depth values causes considerable training difficulty. We divide the depth range of each scene into a constant number of depth bins and the depth regression is reformulated as the combination of the classification of depth bins and the regression of the residual depth values, contributing to the depth learning procedure.

%Focusing on important-aware objects, specifically those nearby with sufficient visual cues, can achieve a more balanced and effective learning. It ensures that the importance-aware objects are prioritized and given greater attention, allowing for a more thorough learning process.

% Regarding geometric priors, we introduce a generic depth sampling strategy accompanied by a modified depth loss. Instead of employing uniform depth sampling, we propose two novel sampling strategies: Logarithmic Increment Distribution (SID) and Linear Increment Distribution (LID). 
%
% These strategies allow us to sample dense points near the camera and sparse points in the distance. The depth sampling strategy affects the geometry priors and necessitates modifications to the depth rendering supervision.
%
% Correspondingly, we modify the original depth loss by employing a Bin Increase loss function. Similar to the sampling strategy, we increase the penalty weight for the predictions of nearby depth value while reducing the penalty weight for the predictions of distant depth value. The Bin Increase loss ensures that the network focuses more on predicting depth values of objects nearby while being less influenced by distant objects.

%Our \algorithmname~integrates geometry and semantic priors in 3D object detection, which uses volume rendering techniques, enabling the 3D volumes to contain both geometry and semantic priors.
%This way, our approach enhancements effectively combine both priors to achieve more accurate and robust 3D object detection from multi-view images. 

The resulting algorithm, termed \algorithmname, is extensively evaluated on the widely used ScanNetV2~\cite{ScanNet} and ARKITScenes~\cite{ARKitScenes} datasets, which are prominent benchmarks for indoor 3D detection. Specifically, our approach achieves $53.9$ mAP$@0.25$ and $29.6$ mAP$@0.5$ in ScanNetV2, surpassing the performance of previous detectors by a large margin. Encouraging results are also observed on the ARKITScenes dataset, further demonstrating the superiority of our method.

Our key contributions are summarized as follows:
\begin{enumerate}
    \item We point out three critical flaws in NeRF-Det, \emph{i.e.}, semantic ambiguity, inappropriate sampling, and insufficient utilization of depth supervision.
    \item We propose systematic solutions to address the aforementioned shortcomings, including designing the Semantic Enhancement module, introducing the Perspective-aware Sampling policy, and putting forward the Ordinal Residual Depth Supervision.
    \item Our \algorithmname~consistently outperforms previous indoor multi-view 3D detectors on the ScanNetV2 and ARKITScenes datasets.
\end{enumerate}

\section{Related work}
\label{sec:relatedwork}
%////////////////////////////////

%\yuenan{We can refer to the related work section of NeRF-Det.}

%\yuenan{We need to include the distinctions between our \algorithmname~and previous algorithms.}

To compare the distinctions between our method and the existing methods, the related works are introduced from three aspects: neural radiance field (NeRF), multi-view 3D detection, and indoor multi-view 3D object detection with NeRF.

\noindent \textbf{Neural Radiance Field (NeRF).}
NeRF~\cite{NeRF} is the pioneering work for novel view synthesis and reconstruction.
%which leverages volume rendering to generate neural scene representation. 
The main idea of NeRF is that the geometry and appearance of a scene can be modeled using a continuous and implicit radiance field parameterized by a Multi-Layer Perceptron (MLP). 
%It predicts the density and color for each position in a single scene.
%Traditional NeRF focuses on novel view synthesis, neglecting to extract high-quality 3D object meshes. Specifically, in object reconstruction, the weight function $\omega(t)$ and the volume density function $\sigma(t)$ in the traditional volume rendering formulation when approaching maximum values are inconsistent. This limitation often leads to suboptimal results with numerous noisy points.
%To further improve the reconstruction quality of NeRF, 
Based on NeRF, NeuS~\cite{NeuS} and VolSDF~\cite{VolSDF} utilize the Signed Distance Function (SDF) to replace the density and achieve better reconstruction quality.
Due to the spectral bias of MLP, it is more inclined to learn low-frequency signals, causing unsatisfactory 3D reconstruction results with volume rendering. NeuralWarp~\cite{NeuralWarp} utilizes a warping-based loss function to extract the color information from alternate viewpoints, improving reconstruction outcomes.
RegSDF~\cite{RegSDF} proposes several practical regularization terms, including using MVS to obtain additional point clouds and Principal Component Analysis (PCA) to obtain normals as geometric constraints.
Mip-NeRF~\cite{Mip-NeRF} introduces a multi-scale representation for Anti-Aliasing NeRF and replaces Positional Encoding (PE) with Integrated Positional Encoding (IPE). 
%It adopts a visual frustum-based sampling strategy, prioritizing sampling points closer to the viewer and reducing sampling density for distant areas. It significantly improves overall performance. 
It employs a visual frustum-based sampling strategy, prioritizing sampling points closer to the viewer and reducing the sampling density for distant regions. This approach leads to substantial improvements in overall performance.
Although our method and Mip-NeRF concentrate on sampling strategies, our approach explicitly modifies the sampling strategy, whereas Mip-NeRF capitalizes on the frustum characteristics.
NeuS2~\cite{NeuS2} and Neuralangelo~\cite{Neuralangelo} use a hash grid to speed up the training of NeuS, remarkably shortening the per-scene optimization duration of NeRF.
% NeuS2 uses multi-resolution hash coding proposed Instant-NGP instead of MLP to speed up training. However, direct use of hash encodings is impossible because they are spatially discontinuous. A practical method for calculating the spatial gradient of SDFNetwork is thus proposed.
% Compared to the efficient calculation method of hash grid analytic gradients proposed by NeuS, Neuralangelo uses numerical gradients directly. The advantage is that analytic gradients are only relevant to the local hash grid. Due to linear interpolation, calculating numerical gradients introduces more hash grids, making the resulting surface smoother.
Since NeRF is notorious for the tedious per-scene optimization, IBRNet~\cite{IBRNet} resolves this problem by integrating IBR, NeRF, and the designed ray Transformer to predict and render the results, where relative perspective is used instead of the absolute perspective of the original adjacent image. 
NeRF and its variants can capture intricate structural details of 3D scenes and facilitate obtaining 3D scene understanding using only RGB images with known poses. Therefore, the representation of NeRF is well-suited for multi-view 3D object detection tasks.

\noindent \textbf{Multi-view 3D Detection.}
Multi-view 3D object detection aims to predict the category and 3D position of objects by taking multi-view images as input. 
One commonly used approach is to incorporate geometric consistency as a constraint. For example, VEDet~\cite{VEDet} leverages 3D multi-view consistency to improve object localization by integrating viewpoint awareness and equal variance, enhancing 3D scene understanding and geometry learning. It employs a query-based transformer architecture to encode 3D scenes, enriching image features by incorporating positional encoding of 3D perspective geometry.
Furthermore, methods such as DETR3D~\cite{DETR3D} and PETR~\cite{PETR} build upon DETR~\cite{DETR} and combine 2D features from multi-views with 3D position information.
Though the preceding approaches utilize geometric priors to some extent to construct the 3D world, the knowledge of the 3D world is inherently encoded within the network. Therefore, another category of methods focuses on constructing alternative representations. For example, ImVoxelNet~\cite{ImVoxelNet} adopts the voxel-based representation and projects 2D features back onto a 3D grid. However, it lacks explicit geometric information.
Frustum3D~\cite{Frustum3D} constructs a 3D volume using RGB-D data and performs 3D detection on the point cloud. While these construction methods offer more explicit representations of the 3D world, they often consume significant memory resources.

\noindent \textbf{Indoor Multi-view 3D Detection with NeRF.} Recent trends favor the incorporation of NeRF into the detector. %where NeRF-RPN~\cite{NeRF-RPN} and NeRF-Det~\cite{NeRF-Det} are two representative work. 
NeRF-RPN~\cite{NeRF-RPN} breaks new ground by incorporating NeRF into indoor multi-view 3D object detection. It introduces a versatile pre-trained NeRF model for 3D object detection that does not rely on class labels. It utilizes a novel voxel representation, integrating multi-scale 3D neural volumetric features, allowing for direct utilization of NeRF without the need for viewpoint rendering. However, NeRF-RPN does not fully exploit the potential of NeRF. The early-stage grid sampling results in the loss of substantial RGB, density, and geometric information, leading to unsatisfactory performance.
NeRF-Det~\cite{NeRF-Det} is a concurrent work to NeRF-RPN. It combines a NeRF branch with a 3D detection branch, using a shared MLP to exchange the geometric information. However, NeRF-Det is limited by semantic ambiguity, an inappropriate sampling strategy, and inadequate depth supervision, which hinders its effectiveness. In contrast, our \algorithmname~overcomes these limitations and substantially improves the performance of indoor 3D detection.

% In comparison, our approach builds upon NeRF-Det, which serves as a more appropriate benchmark. We effectively address the issues of insufficient semantic information and incomplete utilization of depth information in NeRF-Det, resulting in notable improvements in 3D detection performance. Our method presents a refined and comprehensive solution to enhance the capabilities of NeRF-based 3D indoor object detection methods.

\section{Method}
\label{sec:method}
%////////////////////////////////

In this section, we first have a brief retrospection of the NeRF-Det framework, which serves as the footstone of our work. Then, we point out the shortcomings of NeRF-Det and present corresponding solutions to tackle these limitations.

\subsection{Framework Overview}

\begin{figure*}[tbp]
\centering
\includegraphics[width=\textwidth]{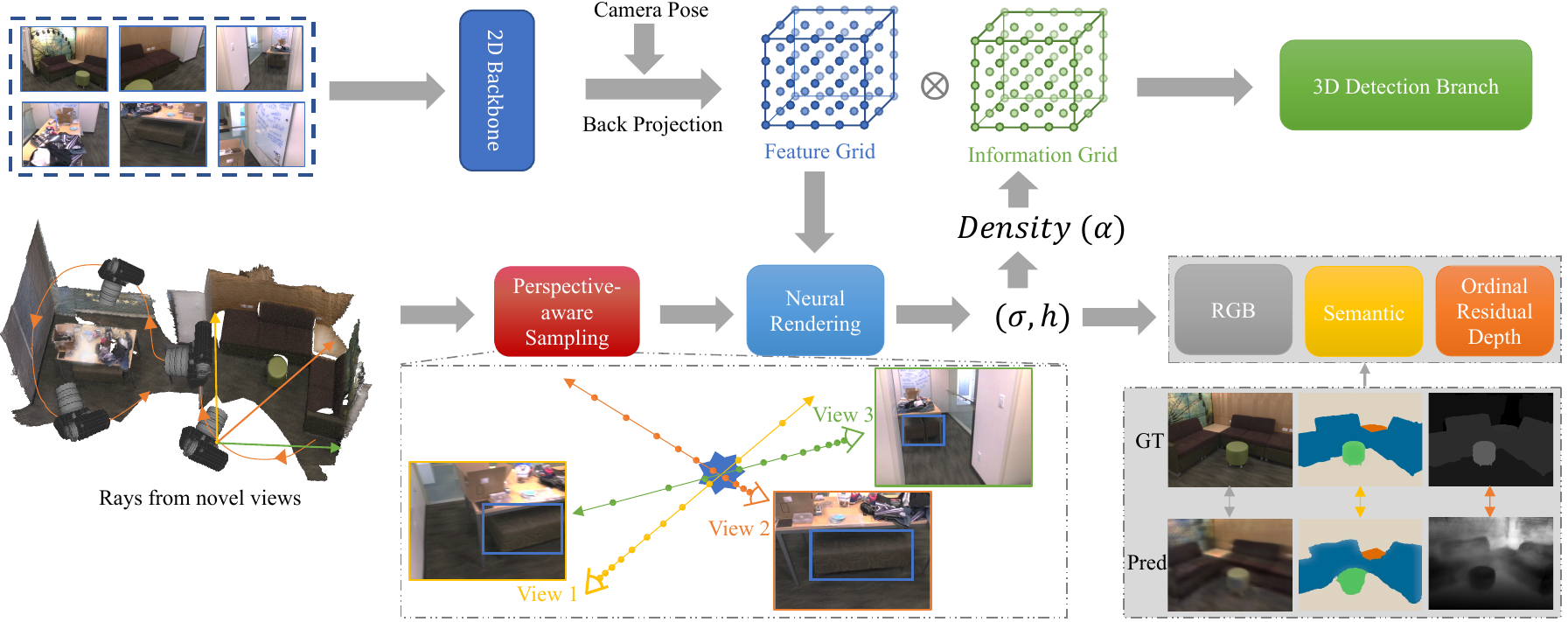}
\caption{\textbf{Schematic overview of our \algorithmname.} 
The framework comprises two branches, i.e., detection and neural rendering.
For the detection side, given the multi-view images, we utilize the 2D backbone to extract discriminative features. The camera pose of each image is used to back-project these 2D features to the 3D space, producing the 3D feature grid. As for the rendering branch, we design the Perspective-aware Sampling policy to concentrate on sampling points of more prominent regions. Two MLPs, \emph{i.e.}, the $\Phi_\text{G}$ and $\Phi_\text{C}$, are employed to estimate the volumetric density and color of the sampled points, respectively. To enhance the semantic awareness of the detection features, we introduce the Semantic Enhancement module $\Phi_\text{S}$, which applies semantic supervision to the 2D views. The $\Phi_\text{G}$ is shared and used to produce the opacity field multiplied by the 3D feature grid to generate the geometry-enhanced features. Ultimately, the reinforced features are fed to the detection head to yield the detection outputs.}
\label{Fig: pipeline}
\end{figure*}

%\yuenan{You can read some relevant papers and mimic the way they tell the story.}

%\yuenan{We need to rename the used modules of NeRF-Det.}

%\yuenan{We need to present the intuitions as well as the generalization of our designs. Merely writing down what we have done is insufficient.}

Given $T$ multi-view images and their pose information, the objective of multi-view 3D detection is to estimate the spatial location and category of the objects, where $T$ represents the total number of views. NeRF-Det introduces a novel approach for indoor multi-view 3D detection by leveraging NeRF to enhance the geometric awareness of the detector. In concrete, it first extracts the 2D features of each posed image from the 2D image backbone. Then, NeRF-Det fuses multi-stage features to generate high-resolution features $\mathbf{F}_{i}\in \mathbb{R}^{C\times H/4\times W/4}$ ($i=1,2,\dots, T$), where $H$ and $W$ denote the height and width of the input image, respectively, and $C$ denotes the number of channels of the feature map. These features are multiplied with the opacity field, estimated by the shared geometry MLP of NeRF to eliminate the feature ambiguity, yielding geometry-aware features.
Despite the excellent performance exhibited by NeRF-Det, it still suffers from three decisive flaws, including semantic ambiguity, inappropriate sampling, and insufficient utilization of depth supervision. To cope with the preceding problems, we proposed the following solutions: \emph{i.e.}, semantic enhancement, perspective-aware sampling, and ordinal residual depth supervision. The schematic overview of our method is shown in Fig.~\ref{Fig: pipeline}.

%Our approach explicitly incorporates high-level semantic and geometric information in the construction of NeRF, enabling the generation of informative 3D volumes from multi-view images. The 3D feature volume encompasses not only low-level RGB and density information but also high-level semantic labels. So, the generated 3D feature volume can contain rich and abundant information, facilitating the subsequent 3D object detection task. 
%Integrating semantic and geometric information aligns well with the classification and localization sub-task in 3D object detection, providing important auxiliary information and improving the performance of 3D object detection. 
%Moreover, addressing the phenomenon that the use of depth supervision in NeRF-Det does not always enhance performance, we propose two improvements, including the sampling strategies and the design of the depth loss.

\subsection{Semantic Enhancement}
Although NeRF-Det achieves impressive performance in image-based indoor detection, it is usually confronted with inaccurate semantic predictions~\cite{Ye2023IntrinsicNeRF} and low confidence scores. The semantic ambiguity problem of NeRF-Det adversely affects the performance of 3D object detection. Therefore, it is necessary to incorporate semantic information into the current framework. 
To this end, we introduce the semantic enhancement module to incorporate high-level semantic information while constructing the 3D volume explicitly. 
Concretely, we integrate the semantic branch $\Phi_\text{S}$ after the geometric module $\Phi_\text{G}$. 
As depicted in Fig.~\ref{Fig: pipeline}, the hidden features $\mathbf{h}(\mathbf{x})$, generated by $\Phi_\text{G}$, are then fed into the $\Phi_\text{S}$ module, producing the semantic predictions:
\begin{equation}
\label{eq:SMLP}
    \mathbf{s} = \mathbf{\Phi_\text{S}(\mathbf{x}, \mathbf{d}, \mathbf{h}(\mathbf{x}))},
\end{equation}
\noindent where $\Phi_\text{G}(.)$ denotes the geometric multi-layer perceptron (MLP), $\mathbf{d}$ is the view direction, $\mathbf{x}$ is the coordinate of the sampled point and $\mathbf{h}(\mathbf{x}) = \mathbf{\Phi_\text{G}}(\mathbf{F}, \mathbf{x})$.
%$\mathbf{s(x)}=\mathbf{\Phi_\text{S}}(\mathbf{x}, \mathbf{d}, \mathbf{\Phi_\text{G}}(\mathbf{x}, \mathbf{d}))$
We use the cross entropy loss to supervise the learning of semantic maps:
\begin{equation}
\label{eq:Segloss}
    \mathcal{L}_{\text{Seg}}(\mathbf{s}, \hat{\mathbf{s}}) = \text{CE}(\mathbf{s}, \hat{\mathbf{s}}).
\end{equation}

\noindent Here, $\hat{\mathbf{s}}$ is the ground-truth semantic label. By incorporating a semantic branch and leveraging the $\Phi_\text{G}$ and $\Phi_\text{S}$ modules, our method enables the 3D feature volume to capture and effectively utilize valuable semantic information, thus increasing the semantic awareness of the multi-view detector. Note that the semantic branch is exclusively utilized during training and imposes no additional burden during testing. Therefore, our approach enhances the detection performance without incurring extra computational costs.

% \subsection{Enhanced Depth Learning}
\subsection{Perspective-aware Depth Supervision}
Depth learning is an important component of the multi-view detector as it serves as the bridge to connect the 2D perspective view and the 3D space. Recall that NeRF-Det assigns the same penalty to all regions for the depth estimation. However, since distant objects with fewer visual cues often lead to higher losses, the depth loss of distant areas will dominate the training process, making the depth learning of nearby regions ineffective. 
% Moreover, one object that is distant in one view can be close in another view. 
Moreover, objects far away in one view can be close in another.
The multi-view complementation indicates that these distant objects can benefit from more accurate depth values from alternative perspectives. Consequently, we propose prioritizing the depth learning on close objects and reducing the weight of distant objects.

To tackle the preceding problems, we propose enhancing the depth learning of the multi-view detector from two perspectives, \emph{i.e.}, the loss function, and the sampling policy. Remember that NeRF-Det directly regresses the ground-truth depth values. However, the direct regression of the continuous depth values is extremely difficult. Instead, we transform the original regression prediction of continuous depth values into the classification prediction of discrete depth bins and the regression of the continuous residual depth values. This practice draws inspiration from the depth sampling policy in monocular 3D detection~\cite{DORN,Center3D}, and we redesign it to suit the indoor 3D detection task better, as presented in Eqn.~\ref{eq:LgIS} and~\ref{eq:LnIS}. Compared to monocular 3D detection, the depth range of indoor scenes is relatively limited. Hence, we divide the depth subtly, and $N$ is the number of divided bins.

%\yuenan{This part needs polishing}

\noindent \textbf{Perspective-aware Sampling}. Considering that distant information is challenging to learn and may also contain inaccuracies, we replace the  Uniformly Sampling (US) of NeRF-Det with the perspective-aware sampling strategy. The original uniform sampling strategy is given as follows:

\begin{equation} %\text{US}_{indoor}:
%\begin{flalign}
\label{eq:US}
\quad l_{\text{US}}  = (N-1)\frac{z-z_{\min}}{z_{\max} - z_{\min}}, 
%\end{flalign}
\end{equation}

\noindent where $z$ denotes the depth, $z_{\max}$ and $z_{\min}$ represents the maximum and minimum depth of each scene.

%If both near and far areas are sampled using the same distribution, the network will be penalized heavily by the distant regions. In addition, since there are images from multiple perspectives, although the object is farther from this perspective, it has another closer perspective. Therefore, prediction cannot be limited to a single perspective. The performance can be improved by concentrating more on nearby easy-to-learn objects and less on distant targets with insufficient visual information. Consequently, the focus will be put on the more difficult distant parts, resulting in sub-optimal learning while neglecting the nearby areas that should be easier to learn.

To fully harness the precious visual clues in the nearby areas, we propose Logarithmic Increment Sampling  (LgIS), which applies logarithmic increments based on the US. The mathematical formulation of LgIS is shown as follows: 
\begin{equation} %\text{LgIS}_{indoor}:
%\begin{flalign}
\label{eq:LgIS}
\quad l_{\text{LgIS}} = (N-1)\frac{\log z-\log z_{min}}{\log z_{\max} - \log z_{\min}}.
%\end{flalign}
\end{equation}
% \noindent where $z$ denotes the depth, and $N$ is the number of bins. 
\noindent This sampling policy allows the network to allocate its learning resources more reasonably and prioritize the parts conducive to initial learning. 
Additionally, we propose Linear Increment Sampling(LnIS), introducing linear increments to guide the attention of the network toward more easily learned parts. The specific formulation is given below:

% \begin{equation}
%\begin{flalign}
%\label{eq:UID}
%\text{UIS}_{indoor}:\quad l & = (N-1)\frac{z_{\max}z_{\min}}{z_{\max} - z_{\min}}(\frac{1}{z_{\min}} - \frac{1}{z}), &
%\end{flalign}
% \end{equation}

\begin{equation} %\text{LnIS}_{indoor}:
%\begin{flalign}
\label{eq:LnIS}
\begin{aligned}
l_{\text{LnIS}} & =-0.5+0.5\sqrt{1+4\delta}, & \\
\text{where} \quad \delta & =N(N-1)\frac{z-z_{\min}}{z_{\max}-z_{\min}}. &
\end{aligned}
%\end{flalign}
\end{equation}

The LgIS and LnIS sampling policies constitute our perspective-aware sampling that prioritizes nearby areas while de-emphasizing distant areas.

\noindent \textbf{Ordinal Residual Depth Supervision}. We can encode an instance depth $z$ in $l_\text{int} = \lfloor l \rfloor$ ordinal bins and the estimated depth is decomposed as $z=z_{l_\text{int}} + z_\text{res}$, where $\lfloor . \rfloor$ is the floor operation.
Furthermore, considering the online learning aspect of the rendering part in NeRF, we have made corresponding adjustments to the depth loss, considering the characteristics of LnIS and LgIS. 
The proposed Ordinal Residual Depth loss comprises the classification of ordinal depth bins and the regression of the residual depth value, as presented in Eqn.~\ref{eq:depthloss}.

\begin{equation}
\label{eq:depthloss}
\begin{aligned}
    \mathcal{L}_{\text{Depth}}(z, \hat z) = \text{CE}(l_{\text{int}}, \hat{l}_{\text{int}}) + \gamma \text{L1}(z_{\text{res}}, \hat{z}_{\text{res}}),
\end{aligned}
\end{equation}
\noindent where 
% $p$ is the probability of $l_{int}$ and
$\gamma$ is the weight of residual depth loss.
%The $L_{Depth_{ori}}(z, \hat z)$ can be chosen by any depth loss, such as L1 loss, L2 loss, etc.

%By introducing these modifications, we aim to alleviate the issues related to depth regression and improve the handling of distant objects in indoor 3D detection. We also use the proposed $\text{LgIS}$ and $\text{LnIS}$ sampling methods to replace the uniform sampling (US) (as shown in Eqn.~\ref{eq:US}) in ray sampling. The proposed $\text{LgIS}$ and $\text{LnIS}$ strategies provide a more effective and appropriate solution for the specific challenges of indoor 3D object detection.\looseness=-1

Besides, following the multi-level geometry reasoning framework in NeRF, we further introduce a fine sub-network that aligns with the structure of the coarse sub-network, allowing for more precise sampling.

\subsection{Training Objective}
Regarding the detection branch, we follow NeRF-Det~\cite{NeRF-Det} and employ the conventional detection loss $\mathcal{L}_{\text{Det}}$, comprised of the center regression loss, the box size regression loss, and the categorical classification loss.
As for the rendering branch, we use a photo-metric loss $\mathcal{L}_{\text{RGB}}$ to learn the low-level appearance information, a geometric loss $\mathcal{L}_{\text{Geo}}$ to learn geometry, and a semantic segmentation loss $\mathcal{L}_{\text{Seg}}$ to learn the high-level semantic information.
When the ground-truth depth is used, the depth estimation loss $\mathcal{L}_\text{Depth}$, computed by Eqn.~\ref{eq:depthloss}, can be further utilized.
The overall training objective is shown in Eqn.~\ref{eq:loss}.

\begin{equation}
\label{eq:loss}
\begin{aligned}
    \mathcal{L} = \mathcal{L}_{\text{Det}} + \mathcal{L}_{\text{RGB}}  + \mathcal{L}_{\text{Geo}} + \mathcal{L}_{\text{Seg}} + \mathcal{L}_{\text{Depth}}.
\end{aligned}
\end{equation}

We empirically find that setting all the loss coefficients to one yields the best performance.
%\yuenan{The loss coefficients are all one?} Answer: Yes.

%&(\mathcal{L}_{\text{centerness}}(\theta) + \mathcal{L}_{\text{bbox}}(\theta) + \mathcal{L}_{\text{classes}}(\theta))

%Even though we use the NeRF branch, modify the geometric branch, and added a semantic branch during training, it is not required in inference. 
%Therefore, our proposed branches can be used in other 3D tasks. Moreover, our trained network is generalizable to new scenes that are never seen during training.\looseness=-1

\section{Experiments}
\label{sec:experiments}
%////////////////////////////////
\begin{table*}[tb]
\centering
\caption{Quantitative comparison on ScanNetV2. The first block shows point-cloud-based and RGBD-based methods; the others are multi-view RGB-only methods. $*$ indicates the method with depth supervision. $\dagger$ means our retest follows the official project. The evaluation of each category is based on mAP$@0.25$. We use \textbf{bold} to indicate our method outperforms other approaches under the same configuration.}
\resizebox{\textwidth}{!}{
\begin{tabular}{@{}lcccccccccccccccccccc@{}}
\toprule
\textbf{Methods} & cab & bed & chair & sofa & table & door & wind & bkshf & pic & cntr & desk & curt & fridge & shower & toil & sink & bath & ofurn & \textbf{mAP$@.25$} & \textbf{mAP$@.50$}\\
\midrule
Seg-Cluster~\cite{SGPN} & 11.8 & 13.5 & 18.9 & 14.6 & 13.8 & 11.1 & 11.5 & 11.7 & 0.0 & 13.7 & 12.2 & 12.4 & 11.2 & 18.0 & 19.5 & 18.9 & 16.4 & 12.2 & 13.4 & -\\
Mask R-CNN~\cite{MaskRCNN} & 15.7 & 15.4 & 16.4 & 16.2 & 14.9 & 12.5 & 11.6 & 11.8 & 19.5 & 13.7 & 14.4 & 14.7 & 21.6 & 18.5 & 25.0 & 24.5 & 24.5 & 16.9 & 17.1 & 10.5\\
SGPN~\cite{SGPN} & 20.7 & 31.5 & 31.6 & 40.6 & 31.9 & 16.6 & 15.3 & 13.6 & 0.0 & 17.4 & 14.1 & 22.2 & 0.0 & 0.0 & 72.9 & 52.4 & 0.0 & 18.6 & 22.2 & -\\
% 3D-SIS~\cite{3DSIS} & 12.8 & 63.1 & 66.0 & 46.3 & 26.9 & 8.0 & 2.8 & 2.3 & 0.0 & 6.9 & 33.3 & 2.5 & 10.4 & 12.2 & 74.5 & 22.9 & 58.7 & 7.1 & 25.4 & -\\
3D-SIS~\cite{3DSIS} & 19.8 & 69.7 & 66.2 & 71.8 & 36.1 & 30.6 & 10.9 & 27.3 & 0.0 & 10.0 & 46.9 & 14.1 & 53.8 & 36.0 & 87.6 & 43.0 & 84.3 & 16.2 & 40.2 & 22.5\\ % (w/ RGB)
VoteNet~\cite{VoteNet} & 36.3 & 87.9 & 88.7 & 89.6 & 58.8 & 47.3 & 38.1 & 44.6 & 7.8 & 56.1 & 71.7 & 47.2 & 45.4 & 57.1 & 94.9 & 54.7 & 92.1 & 37.2 & 58.7& 33.5\\
FCAF3D~\cite{fcaf3d} & 57.2 & 87.0 & 95.0 & 92.3 & 70.3 & 61.1 & 60.2 & 64.5 & 29.9 & 64.3 & 71.5 & 60.1 & 52.4 & 83.9 & 99.9 & 84.7 & 86.6 & 65.4 & 71.5 & 57.3\\
CAGroup3D~\cite{CAGroup} & 60.4 & 93.0 & 95.3 & 92.3 & 69.9 & 67.9 & 63.6 & 67.3 & 40.7 & 77.0 & 83.9 & 69.4 & 65.7 & 73.0 & 100 & 79.7 & 87.0 & 66.1 & 75.12 & 61.3\\
\hline
ImVoxelNet-R50 & 34.5 & 83.6 & 72.6 & 71.6 & 54.2 & 30.3 & 14.8 & 42.6 & 0.8 & 40.8 & 65.3 & 18.3 & 52.2 & 40.9 & 90.4 & 53.3 & 74.9 & 33.1 & 48.4 & 23.7\\
NeRF-Det-R50 & 37.2 & 84.8 & 75.0 & 75.6 & 51.4 & 31.8 & 20.0 & 40.3 & 0.1 & 51.4 & 69.1 & 29.2 & 58.1 & 61.4 & 91.5 & 47.8 & 75.1 & 33.6 & 52.0 & 26.1\\
\rowcolor{lightgray}\algorithmname-R50(Ours) & 34.8 & \textbf{86.3} & \textbf{75.7} & \textbf{79.9} & \textbf{56.9} & \textbf{34.3} & \textbf{25.4} & \textbf{53.3} & \textbf{2.3} & \textbf{51.8} & \textbf{70.5} & \textbf{32.6} & \textbf{54.3} & 51.3 & \textbf{92} & \textbf{53.3} & \textbf{80.7} & \textbf{34.1} & \textbf{53.9(+1.9)} & \textbf{29.6(+3.5)}\\
NeRF-Det-R50$*$ & 37.7 & 84.1 & 74.5 & 71.8 & 54.2 & 34.2 & 17.4 & 51.6 & 0.1 & 54.2 & 71.3 & 16.7 & 54.5 & 55.0 & 92.1 & 50.7 & 73.8 & 34.1 & 51.8 & 27.4\\
\rowcolor{lightgray}\algorithmname-R50$*$(Ours) & \textbf{38.9} & 83.9 & 73.9 & \textbf{77.6} & \textbf{57.2} & 33.3 & \textbf{23.5} & 47.9 & \textbf{1.51} & \textbf{56.9} & \textbf{77.7} & \textbf{21.1} & \textbf{61.5} & 46.8 & \textbf{92.8} & 49.2 & \textbf{80.2} & \textbf{34.5} & \textbf{53.2(+1.4)} & \textbf{29.6(+2.2)}\\
ImVoxelNet-R101 & 30.9 & 84.0 & 77.5 & 73.3 & 56.7 & 35.1 & 18.6 & 47.5 & 0.0 & 44.4 & 65.5 & 19.6 & 58.2 & 32.8 & 92.3 & 40.1 & 77.6 & 28.0 & 49.0 & -\\
NeRF-Det-R101 & 36.8 & 85.0 & 77.0 & 73.5 & 56.9 & 36.7 & 14.3 & 48.1 & 0.8 & 49.7 & 68.3 & 23.5 & 54.0 & 60.0 & 96.5 & 49.3 & 78.4 & 38.4 & 52.9 & -\\
\rowcolor{lightgray}\algorithmname-R101(Ours) & 36.1 & 82.9 & 74.9 & \textbf{79.1} & \textbf{57.0} & \textbf{37.3} & \textbf{24.9} & \textbf{54.6} & \textbf{2.4} & \textbf{51.7} & \textbf{72.2} & \textbf{25.5} & \textbf{58.7} & 51.5 & 92.7 & \textbf{50.8} & \textbf{82.2} & 35.1 & \textbf{53.9(+1.0)} & \textbf{30.0}\\
NeRF-Det-R101$*$ & 37.6 & 84.9 & 76.2 & 76.7 & 57.5 & 36.4 & 17.8 & 47 & 2.5 & 49.2 & 52 & 29.2 & 68.2 & 49.3 & 97.1 & 57.6 & 83.6 & 35.9 & 53.3 & -\\
NeRF-Det-R101$*^\dagger$ & 38.7 & 81.3 & 75.7 & 76.8 & 58.1 & 33.5 & 23.7 & 42.1 & 4.5 & 56.6 & 68.2 & 26.6 & 51.5 & 47.4 & 85.3 & 49.3 & 73.1 & 30.8 & 51.3 & 27.4\\ % test twice, can not get the 52.6 mAP, which is report in his GitHub
\rowcolor{lightgray}\algorithmname-R101$*$(Ours) & 38.7 & \textbf{85.0} & 73.2 & \textbf{78.1} & 56.3 & \textbf{35.1} & 22.6 & \textbf{45.5} & 1.9 & 50.7 & \textbf{72.6} & 26.5 & \textbf{59.4} & \textbf{55.0} & \textbf{93.1} & \textbf{49.7} & \textbf{81.6} & \textbf{34.1} & \textbf{53.3(+2.0)} & \textbf{30.0(+2.6)}\\
\bottomrule
\end{tabular}}
\label{Tab: compare}
\end{table*}

\noindent \textbf{Dataset.}
We perform experiments in two popular indoor 3D detection datasets, \emph{i.e.}, ScanNetV2~\cite{ScanNet} and ARKITScenes~\cite{ARKitScenes}. ScanNetV2 contains $1,513$ complex indoor scenes with approximately $2.5 M$ RGB-D frames and is annotated with semantic and instance segmentation for $18$ object categories. Since ScanNetV2 does not provide amodal or oriented bounding box annotation, we follow NeRF-Det~\cite{NeRF-Det} and predict axis-aligned bounding boxes instead. The input image resolution is $320 \times 240$. We mainly evaluate the methods by mAP with $0.25$ IoU and $0.5$ IoU threshold, denoted by mAP$@.25$ and mAP$@.50$.
ARKITScenes contains around $1.6$ K rooms with more than $5,000$ scans. Each scan includes a series of RGB-D posed images. Since some labels about the sky direction of each video are inaccurate~\cite{PARQ}, we rectify the images according to the metadata and use the videos with all sky directions except ``Left''. On ARKITScenes, the input image resolution is $256 \times 192$, and the image feature map size is $64 \times 48$.
% Following NeRF-Det, we utilize the subset of the dataset with low-resolution images. The subset contains $2,257$ scans of $841$ unique scenes, and the image resolution is $256 \times 192$.
We follow the official dataset partition and adopt mAP$@.25$ and mAP$@.50$ as the evaluation metric.

\subsection{Quantitative results}
Table~\ref{Tab: compare} summarizes the comparison between our \algorithmname~and state-of-the-art methods on the validation set of ScanNetV2. Without bells and whistles, our method outperforms all multi-view 3D detectors regarding mAP$@.25$ and mAP$@.50$. For instance, our \algorithmname~significantly outperforms NeRF-Det with ResNet-50 by $+1.9\%$ and $+3.5 \%$ in terms of mAP$@.25$ and mAP$@.50$, respectively.
We also provide experimental results on the ARKITScenes dataset, as presented in Table~\ref{Tab: ARK}. The performance gap between our \algorithmname~and NeRF-Det is $+0.6\%$ in mAP$@.25$ and $+0.9\%$ in mAP$@.50$.
%The encouraging result further proves the effectiveness and generality of our method across diverse datasets and real-world scenarios. 
The consistent improvements observed in two popular indoor 3D detection benchmarks demonstrate the robustness and versatility of our approach.

\begin{table}[tb]
\centering
\caption{Comparisons on ARKITScenes val set.}
% \resizebox{\textwidth}{!}{
\begin{tabular}{@{}lccc@{}}
\toprule
\textbf{Methods} & \textbf{scene} & \textbf{mAP$@.25$} & \textbf{mAP$@.50$}\\
\midrule
ImVoxelNet-R50 & whole-scene & 23.6 & -\\
\hline
\multirow{2}{*}{NeRF-Det-R50} & whole-scene & 26.7 & -\\
 & except ``Left'' & 42.7 & 26.2\\
\hline
\algorithmname-R50 & except ``Left'' & \textbf{43.3(+0.6)} & \textbf{27.1+(0.9)}\\
\bottomrule
\end{tabular}
% }
\label{Tab: ARK}
\end{table}

\subsection{Ablation study}
% We conduct ablation studies on ScanNetV2 and examine the effectiveness of each component in our method. The chosen baseline is NeRF-Det with ResNet-50. For each component, we verify its efficacy with and without depth supervision. To ensure a fair comparison, we keep the experimental settings of NeRF-Det. \looseness=-1

We perform comprehensive ablation studies on the ScanNetV2 dataset to evaluate the effectiveness of each component in our method. Our chosen baseline is NeRF-Det with ResNet-50. We assess the efficacy of each component, both with and without depth supervision. To ensure a fair and consistent comparison, we maintain the experimental settings of NeRF-Det throughout the evaluation process.
%To control variables, we fix the seed to 0, and other training strategies remain consistent with NeRF-Det.

% \noindent \textbf{Effect of the geometric branch.}
% To study the impact of the new geometric branch, we investigate it through extra experiments, as shown in \text{Table~\ref{Tab: geometric}}. 
% The results show that the geometric cues, including SDF, gradient, and normal, can help 3D detection. 
% As components gradually increase, the final accuracy also increases accordingly.
% We can see that the initial performance ($mAP@0.25$ and $mAP@0.5$) is boosted from 
% $52.0\%/26.1\%$ to ???
% without depth supervision, which is rather impressive.
% Moreover, both without depth supervision and with depth supervision work well, suggesting that the proposed geometric branch is robust.
% \begin{table}[tb]
% \centering
% \caption{Ablation study of whether adding geometric branch.}
% \begin{tabular}{@{}lcc@{}}
% \toprule
% \textbf{Methods} & \textbf{mAP$@.25$} & \textbf{mAP$@.50$}\\
% \midrule
% NeRF-Det-R50 & 52.0 & 26.1\\
% + geometric branch & \\
% + SSIM loss & \\
% both & \\
% Improvement & \\
% \hline
% NeRF-Det-R50* & 51.8 & 27.4\\
% + geometric branch & \\
% + SSIM loss & \\
% both & \\
% Improvement & \\
% \bottomrule
% \end{tabular}
% \label{Tab: geometric}
% \end{table}

\noindent \textbf{Effect of Semantic Enhancement Module.}
% The performance is shown in Table~\ref{Tab: semantic}.
% The results show that semantic cues can make the performance of 3D detection better.
% With the Semantic Enhanced module, the 3D feature volume has more high-level semantic information.
% Sufficient feature information makes 3D detection easier and reduces the number of erroneous examples due to inaccurate categories in NeRF-Det.
The performances are presented in Table~\ref{Tab: semantic}, indicating that including semantic cues improves the performance of 3D object detection. 
The semantic enhanced module enriches the 3D feature volume with high-level semantic information. 
This increased availability of feature information facilitates more accurate 3D detection and mitigates the occurrence of erroneous examples caused by inaccurate categorization in NeRF-Det.

\begin{table}[tb]
\centering
\caption{Ablation study of Semantic Enhancement module.}
\resizebox{\linewidth}{!}{
\begin{tabular}{@{}lcccc@{}}
\toprule
\multirow{2}{*}{\textbf{Methods}} & \multicolumn{2}{c}{w/o depth} & \multicolumn{2}{c}{w/ depth}\\
& \textbf{mAP$@.25$} & \textbf{mAP$@.50$} & \textbf{mAP$@.25$} & \textbf{mAP$@.50$}\\
\midrule
NeRF-Det-R50 & 52.0 & 26.1 & 51.8 & 27.4\\
+ Semantic Enhancement & \textbf{53.4(+1.4)} & \textbf{27.9(+1.8)} & \textbf{52.3(+0.5)} & \textbf{28.3(+0.9)}\\
\bottomrule
\end{tabular}}
\label{Tab: semantic}
\end{table}

\noindent \textbf{Effect of Sampling Strategy and Depth Loss.}
We investigate different sampling strategies and depth losses, as shown in Tables~\ref{Tab: sample} and~\ref{Tab: depthloss}. It shows that LgIS and LnIS, which have larger bin sizes in more significant depths, perform better. Additionally, we test inverse depth using uniform sampling (UIS). Although its bin size is increased, the performance is reduced. 
Due to the squared relationship between increasing bin size and depth, UIS ignores distance information.

\begin{table}[tb]
\centering
\caption{Ablation study of sampling strategies.}
\resizebox{\linewidth}{!}{
\begin{tabular}{@{}lcccc@{}}
\toprule
\multirow{2}{*}{\textbf{Methods}} & \multicolumn{2}{c}{w/o depth} & \multicolumn{2}{c}{w/ depth}\\
& \textbf{mAP$@.25$} & \textbf{mAP$@.50$} & \textbf{mAP$@.25$} & \textbf{mAP$@.50$}\\
\midrule
US & 52.0 & 26.1& 51.8 & 27.4\\
UIS & 51.8(-0.2) & 25.8(-0.3)& 51.7(-0.1) & 26.4(-1.0)\\
LgIS & \textbf{53.2(+1.2)} & \textbf{26.3(+0.2)}& \textbf{52.4(+0.6)} & \textbf{28.8(+1.4)}\\
LnIS & \textbf{53.0(+1.0)} & \textbf{27.1(+1.0)}& \textbf{53.2(+1.4)} & \textbf{29.4(+2.0)}\\
\bottomrule
\end{tabular}}
\label{Tab: sample}
\end{table}

For depth loss, we investigate three losses: L1 loss, Huber loss, and the proposed Ordinal Residual Depth loss.
Huber loss combines the benefits of L1 and L2 Loss; the incrementally scaled L1 region reduces sensitivity to outliers, while the L2 region provides smoothness. 
% That is, long-distance uses L1 loss to reduce the impact of depth loss values, while short-distance uses L2 loss to make depth loss values smoother and more detailed.
Consequently, we utilize L1 loss to minimize the impact of depth loss in long-distance while employing L2 loss for short-distance to enhance smoothness and capture finer details.
The Ordinal Residual Depth loss we design discretizes depth and divides it into a bin classification problem and a residual regression problem, simplifying depth learning.
Additionally, based on perceptive-aware sampling, Ordinal Residual Depth loss prioritizes nearby objects. 
The experimental results demonstrate that our proposed Ordinal Residual Depth loss enhances depth learning, improving 3D detection performance.

\begin{table}[tb]
\centering
\caption{Ablation study of depth loss.}
\begin{tabular}{@{}lcc@{}}
\toprule
\textbf{Methods} & \textbf{mAP$@.25$} & \textbf{mAP$@.50$}\\
\midrule
L1 loss & 51.8 & 27.4\\
Huber loss & \textbf{52.9(+1.1)} & 27.1(-0.3) \\
Ordinal Residual Depth loss & \textbf{52.8(+1.0)} & \textbf{27.5(+0.1)}\\
\bottomrule
\end{tabular}
\label{Tab: depthloss}
\end{table}

\begin{table}[tb]
\centering
\caption{Ablation study of the fine sub-network.}
\resizebox{\linewidth}{!}{
\begin{tabular}{@{}lcccc@{}}
\toprule
\multirow{2}{*}{\textbf{Methods}} & \multicolumn{2}{c}{w/o depth} & \multicolumn{2}{c}{w/ depth}\\
& \textbf{mAP$@.25$} & \textbf{mAP$@.50$} & \textbf{mAP$@.25$} & \textbf{mAP$@.50$}\\
\midrule
NeRF-Det-R50 & 52.0 & 26.1 & 51.8 & 27.4\\
+ fine sub-network & \textbf{52.6(+0.6)} & \textbf{27.4(+1.3)} & \textbf{52.0(+0.2)} & \textbf{28.7(+1.3)}\\
\bottomrule
\end{tabular}}
\label{Tab: fine}
\end{table}

\noindent \textbf{Effect of Fine Sub-network.}
Building upon NeRF and its variants, we incorporate a fine sub-network into our approach. The structure of the fine sub-network mirrors that of the coarse sub-network. It performs a finer sampling within the attention area identified by the coarse sub-network. Our experimental results demonstrate that including the fine sub-network yields significant performance improvements.

\noindent \textbf{Effect of Relative Depth Map.}
We study whether relative depth will affect performance, as shown in Table~\ref{Tab: depthnorm}. The experiment shows that relative depth can improve performance significantly, over $1.1\%$ in mAP$@.25$ and $0.2\%$ in mAP$@.50$ compared to absolute depth.
As depth prediction is a complex problem, we convert a prediction of absolute depth into relative depth, simplifying the problem.

\begin{table}[tb]
\centering
\caption{Ablation study of depth normalization.}
\begin{tabular}{@{}lcc@{}}
\toprule
\textbf{Methods} & \textbf{mAP$@.25$} & \textbf{mAP$@.50$}\\
\midrule
NeRF-Det-R50* & 51.8 & 27.4\\
+ depth normalization & \textbf{52.9(+1.1)} & \textbf{27.6(+0.2)}\\
\bottomrule
\end{tabular}
\label{Tab: depthnorm}
\end{table}

\twocolumn[{%
% \vspace{-3em}
    \renewcommand\twocolumn[1][]{#1}%
    \setlength{\tabcolsep}{0.0mm} %0
    \newcommand{\sz}{0.125}  % 0.125 0.11
    \begin{center}
        \newcommand{\teaserwidth}{\textwidth}
    % \vspace{-4em}
        \includegraphics[width=0.99\linewidth]{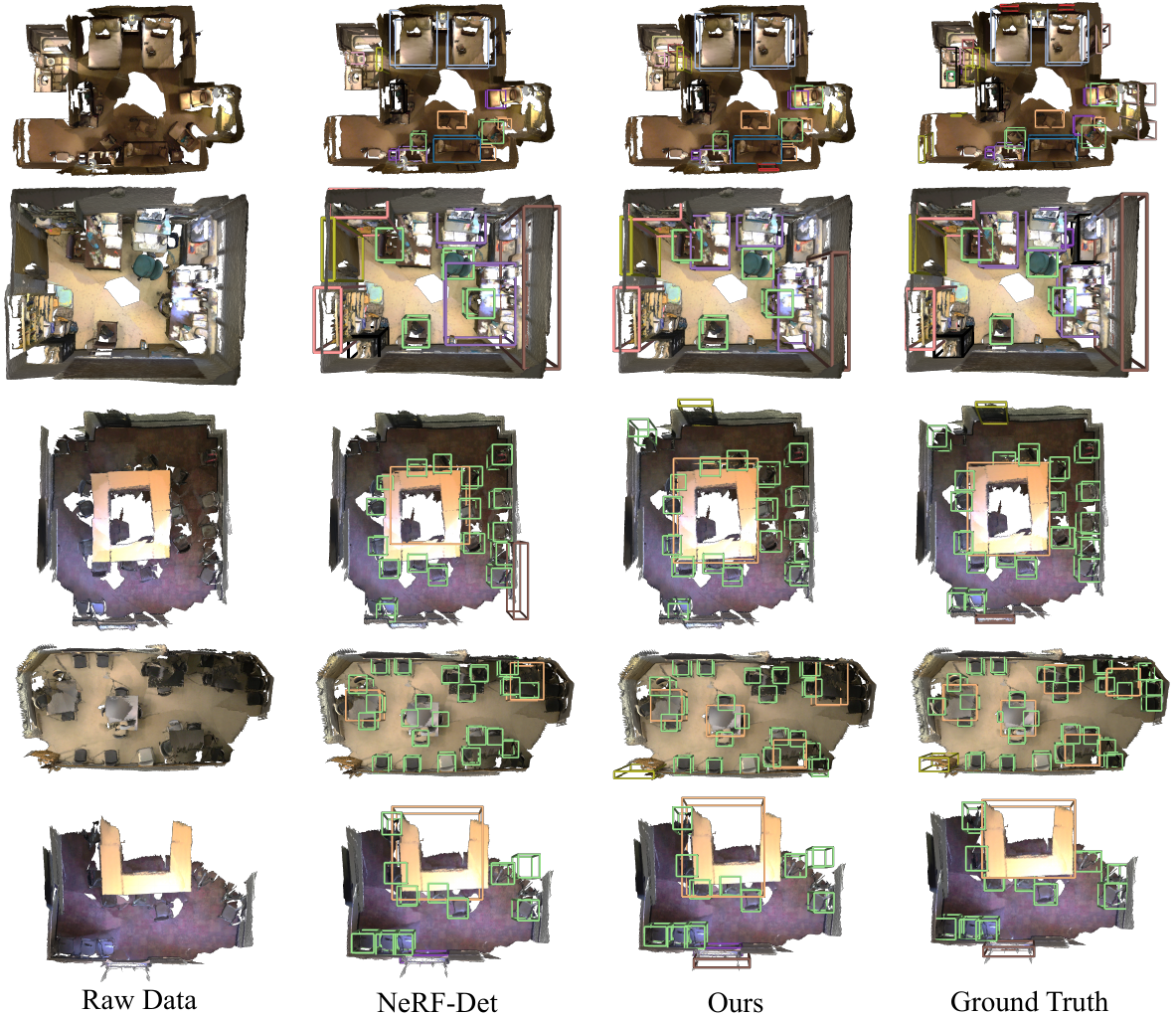}
      % \vspace{-1em}
        \captionof{figure}{\textbf{Visual comparison between NeRF-Det and \algorithmname.} Note that our approach only takes posed RGB images as input. The reconstructed mesh is only used for visualization.}
    \label{Fig: detectionresuts}
    \end{center}
}]

% \begin{figure*}[tbhp]
% \centering
% \includegraphics[width=\linewidth]{figs/vis3.pdf}
% \caption{Qualitative results of the predicted 3D bounding box on \algorithmname. Note that our approach only considers posed RGB images as input; the reconstructed mesh is only for visualization.}
% \label{Fig: detectionresuts}
% \end{figure*}

\subsection{Qualitative results}
% We visualize the prediction of \algorithmname~in Fig.~\ref{Fig: detectionresuts}, which presents some of the challenging cases in NeRF-Det, and our work successfully addresses them.
% We visualize the prediction of NeRF-Det and ours \algorithmname~ under challenging scenes, in Fig.~\ref{Fig: detectionresuts}. Our method demonstrates remarkable accuracy even in challenging scenes, including dense scenes, severe occlusions, significant morphological diversity, etc. 
% The first challenge is the bounding boxes are missing due to factors such as high object density (in the first scene), object occlusion, and embedding relationships (in the fourth scene).
% The second challenge is incorrect object dimensions, as the desk in the second scene exemplifies. 
% The third challenge is positional deviations, such as the table in the third scene. 
% Moreover, the fourth challenge is misclassification, as shown in the fifth scene, which treats the curtain as the garbage bin.
% Lastly, the fifth challenge is false positives, such as adding the "curtain" category in the third scene.
We present visualizations of the predictions made by NeRF-Det and our enhanced \algorithmname, as illustrated in Fig.~\ref{Fig: detectionresuts}. Our method effectively handles diverse challenges, including scenes with high object density, severe occlusion, and significant morphological diversity. Notably, we rectify specific incorrect predictions made by NeRF-Det.
For instance, we successfully resolved the issue of missing bounding boxes in the first, second, and fourth scenes.
Furthermore, we overcome misclassification in the fifth scene, where the ``curtain'' is mistakenly identified as the ``garbage bin''. 
Additionally, we favorably alleviate incorrect object dimensions and positional deviations observed in the second and third scenes, respectively. 
Lastly, in the third scene, our enhanced NeRF-Det++ avoids the false positive of including the ``curtain'' object, which NeRF-Det mistakenly predicts.

% We also provide novel view synthesis and depth estimation visualization results, as shown in Fig. ???. The results are from the test set of ScanNetV2. The proposed method generalizes well on test scenarios.
% Remarkably, extraordinary results have been achieved under relatively difficult circumstances. For example, ....

% \begin{figure*}[tbp]
% \centering
% %\includegraphics[]{./}
% \caption{Qualitative results on new perspective reconstruction on \algorithmname. For each triplet, the left image is the synthetic result, the middle is the ground truth RGB image, and the right image is the estimated depth map.}
% \label{Fig: constructionresuts}
% \end{figure*}

% \begin{figure}[tbph]
% \centering
% \includegraphics[width=\linewidth]{figs/class.png}
% \caption{\textbf{The number of predicted bounding boxes and the performance (mAP$@50$) with changed score threshold.}}
% \label{Fig: comparewithnum}
% \end{figure}

\section{Conclusion}
\label{sec:conclusion}

In conclusion, this paper presents \algorithmname, a novel approach for indoor 3D detection from multi-view images. We identify and address three critical flaws in NeRF-Det. Firstly, to tackle semantic ambiguity, we introduce the Semantic Enhancement module that utilizes semantic supervision for improved classification. Secondly, to address inappropriate sampling, we prioritize nearby objects and leverage the characteristics of multi-views through the design of Perspective-aware Sampling. Lastly, we tackle the issue of insufficient utilization of depth supervision by proposing Ordinal Residual Depth Supervision, which incorporates classification of ordinal depth bins and regression of residual depth values. Extensive experiments conducted on the ScanNetV2 and ARKITScenes validate the superiority of our \algorithmname.

\section{Acknowledge}
\label{sec:acknowledge}
%////////////////////////////////
This work was done during his internship at Shanghai Artificial Intelligence Laboratory. This work is partially supported by the National Key R\&D Program of China(NO.2022ZD0160101).

%% The file named.bst is a bibliography style file for BibTeX 0.99c
\bibliographystyle{named}
\bibliography{ijcai23}

\end{document}